%% file: AAAI_Submission.tex
\documentclass[letterpaper]{article} 
\usepackage{aaai2026}  
\usepackage{times}  
\usepackage{helvet}  
\usepackage{courier}  
\usepackage[hyphens]{url}  
\usepackage{graphicx} 
\usepackage{booktabs}
\usepackage{xr}
\usepackage{natbib}  
\usepackage{caption} 
\usepackage{algorithm}
\usepackage{algorithmic}

\usepackage{newfloat}
\usepackage{listings}
\DeclareCaptionStyle{ruled}{labelfont=normalfont,labelsep=colon,strut=off} 
\floatstyle{ruled}
\newfloat{listing}{tb}{lst}{}
\floatname{listing}{Listing}
\title{Where Norms and References Collide: Evaluating LLMs on Normative Reasoning}
\author{
    Mitchell Abrams\textsuperscript{\rm 1},
    Kaveh Eskandari Miandoab\textsuperscript{\rm 1},
    Felix Gervits\textsuperscript{\rm 2},
    Vasanth Sarathy\textsuperscript{\rm 1},
    Matthias Scheutz\textsuperscript{\rm 1}
}
\affiliations{
    \textsuperscript{\rm 1}Tufts University\\
    \textsuperscript{\rm 2}DEVCOM Army Research Laboratory

    Mitchell.Abrams, Kaveh.Eskandari\_Miandoab, Vasanth.Sarathy, Matthias.Scheutz\{@tufts.edu\} \\
    Felix.Gervits.civ@army.mil
}

\usepackage{bibentry}

\begin{document}

\maketitle

\begin{abstract}
Embodied agents, such as robots, will need to interact in situated environments where successful communication often depends on reasoning over social norms: shared expectations that constrain what actions are appropriate in context. A key capability in such settings is norm-based reference resolution (NBRR), where interpreting referential expressions requires inferring implicit normative expectations grounded in physical and social context. Yet it remains unclear whether Large Language Models (LLMs) can support this kind of reasoning. In this work, we introduce SNIC (\textit{Situated Norms in Context}), a human-validated diagnostic testbed designed to probe how well state-of-the-art LLMs can extract and utilize normative principles relevant to NBRR. SNIC emphasizes physically grounded norms that arise in everyday tasks such as cleaning, tidying, and serving. Across a range of controlled evaluations, we find that even the strongest LLMs struggle to consistently identify and apply social norms—particularly when norms are implicit, underspecified, or in conflict. These findings reveal a blind spot in current LLMs and highlight a key challenge for deploying language-based systems in socially situated, embodied settings.

\end{abstract}


%
\begin{links}
    \link{Code and Dataset}{https://github.com/TheSittingCat/SNIC}
\end{links}

\section{Introduction}
\label{introduction}

Reference resolution is a critical aspect of dialogue processing that contributes to natural language understanding by linking language to real-world entities. This process is not always straightforward or sufficiently achievable through language alone. It requires interpreting language within a situated context, often consisting of social norms that are activated by objects in a scene, tasks, locations, or social roles. Prior work has shown that social norms, irrespective of language, can influence the interpretation of a referent \cite{abrams2022social}. 

Consider an example where a human asks an assistant in a communal kitchen: Human: \textit{“Can you pass me a mug?”} Assistant: \textit{“Okay.”} There are three mugs nearby—one clean and two with leftover coffee. Without additional linguistic cues, the request may seem ambiguous. However, given the context, the intended referent is likely the clean mug because, \textit{normatively}, people avoid drinking from used mugs that are not their own or dirty. The underlying norm applied to this situation is: \textit{you should use clean items when drinking or serving beverages}. If the context instead involved cleaning up after an event, then the request might be understood as referring to the dirty mugs. Without normative reasoning, this interpretation could be lost.

Embodied agents or AI systems must incorporate social norms into reference resolution as they operate in social environments. In interactive situations, \textit{social norms} serve as powerful constraints because they consist of shared expectations that people rely on to coordinate behavior. Norms function similarly to language in that they provide a type of social grammar \cite{bicchieri2005grammar}, guiding interlocutors in resolving referential ambiguity. However, normative reasoning for reference resolution is a challenging task; norms are highly contextual and range from abstract (e.g., \textit{show respect}) to very specific and physically grounded (e.g., \textit{you should pick up your glass by the stairs so nobody knocks it over}). Norms also vary in how strongly they should be followed or violated\footnote{referred to as \textit{deontic force}}, which is particularly relevant in norm conflicts where some norms may need to be violated to satisfy stronger norms \citep{malle2019requirements}.

For reference resolution, norms must be identified from both language and context, then reasoned with to resolve a referent. Large Language Models (LLMs) provide a promising solution for norm identification and reasoning since they implicitly encode world knowledge and perform well on question-answering tasks \citep{brown2020language}. However, norms may be difficult to learn through exposure to large amounts of text, as they are often implicit rather than explicitly stated. It is an open question whether LLMs can perform reference resolution that requires normative reasoning. 


To investigate this challenge, we introduce SNIC, a human-validated diagnostic testbed designed to represent the prescriptive and prohibitive social norms that guide referential interpretation in everyday contexts. Unlike prior norm datasets that focus on abstract moral judgments \citep{forbes2020social}, SNIC emphasizes physically grounded norms embedded in real-world tasks that humans and agents would engage in (e.g., cleaning, serving, tidying). Rather than proposing SNIC as a definitive benchmark, we use it as a controlled environment to probe how LLMs handle normative ambiguity, prioritize between conflicting norms, and respond to implicit expectations when resolving ambiguous referential expressions. Across a series of evaluations, we find that models exhibit inconsistencies; not only across different scenarios but even within near-identical scenes under slightly varied conditions. These findings expose a critical but unexplored weakness in contemporary language models: their ability to consistently reason about physically situated social norms.





\section{Background}

Social norms function as a form of social grammar, shaping expectations for behavior in different contexts \citep{bicchieri2005grammar}. This is because social norms consist of both \textit{empirical expectations} (beliefs about what others do) and \textit{normative expectations} (beliefs about what others expect us to do) \citep{malle2019requirements}. Their power to guide behavior has been attractive to the AI community.


Normative reasoning has been studied in AI and the NMAS (normative multi-agent systems) community through formal representations that encode norms using \textit{deontic logic} and contextual activation conditions \citep{malle2019requirements, sarathy2017mental, malle2020general}. These studies have explored how norms are triggered by environmental and linguistic cues, but their application to reference resolution remains unexplored.



To our knowledge, no existing dataset explicitly evaluates \textit{norm-based reference resolution} (NBRR). There has been NLP work in anaphora and coreference resolution that requires external commonsense knowledge for interpretation: reference resolution tasks, such as the \textit{Winograd Schema Challenge} \citep{levesque2012winograd}, \textit{Winogrande} \citep{sakaguchi2021winogrande} (a reference resolution task that requires common-sense knowledge to resolve ambiguous references), and \textit{CommonsenseQA} \citep{talmor2018commonsenseqa}, test common-sense reasoning but do not isolate normative constraints. We follow this tradition and extend the idea that reference and co-reference resolution require not only commonsense knowledge, but even normative knowledge. We also treat text-based vignettes as a reasonable proxy for physical scenarios, as in prior work like TextWorld \citep{cote2019textworld}, though our focus remains on evaluating norms in text form.




In social norm modeling, NLP systems such as \textit{Delphi} and the \textit{Neural Norm Transformer} have been trained on datasets like \textit{Social Chemistry 101} \citep{forbes2020social}. However, these models primarily capture norms related to moral judgments and social etiquette, lacking the physical and situational grounding needed for norm-based reference resolution. This gap motivates the development of new testbeds and evaluation methods tailored to NBRR
This dataset, for instance, consists of situations from online forums (e.g., \textit{"Asking my boyfriend to stop being friends with his ex"}) and associated moral rules of thumb (e.g., \textit{"It's okay to ask your significant other to stop doing something you're uncomfortable with"}) This gap motivates the development of new benchmarks and evaluation methods tailored to NBRR.


LLMs offer a potential solution for addressing the challenges of situated and norm-based reference resolution as they can capture higher-level contextual information and world knowledge and perform well on question-answering tasks and common-sense reasoning \cite{brown2020language}.  \citet{he2024norm} evaluated LLMs on detecting norm violations in household task sequences, using textual scenarios with human-annotated ground truth. Their task focused on post-hoc classification of whether an agent’s actions violated a given norm, with results ranging from 62.5\% to 100\% accuracy across generic and role-based norms using ChatGPT-4. This suggests that LLMs can vary in the types of norms they can detect. Our work, in contrast, addresses a distinct challenge: resolving ambiguous referring expressions where the correct interpretation hinges on proactively applying implicit or conflicting social norms.


There is still other work that highlights LLMs and general social reasoning. \citet{bian2023chatgpt} shows LLMs struggle more with social common sense reasoning with the SocialIQA dataset \citep{sap2019socialiqa} and social factors of language outside of content \citep{hovy2021importance}, in addition to tasks involving implicit social norms and physical grounding \citep{bian2023chatgpt, sap2019socialiqa}. Other work has looked the LLM evaluations of moral dilemmas which are related to social norms \citep{sachdeva2025normative}. 





The integration of LLMs into robotics and embodied AI suggests potential applications for norm-based reference resolution. In robot dialogue systems, LLMs have been used for \textit{object disambiguation} \citep{jiang2024llms}, \textit{command disambiguation} \citep{park2023clara}, and \textit{household organization tasks} \citep{wu2023tidybot}. However, even in these applications, LLMs struggle with handling norm-relevant distinctions. For example, \citet{wu2023tidybot} found that when generating user preference rules for tidying, LLMs occasionally overgeneralized norms, leading to incorrect object groupings (e.g., treating all drawers as equivalent despite cultural differences in organization). These limitations highlight why NBRR remains an open problem: while LLMs can encode broad world knowledge, it is unclear whether they reliably identify and apply context-specific norms for NLP tasks like reference resolution.

Existing datasets in reference resolution and social norm modeling fail to integrate the physical and situational constraints required for resolving referents in a normative context. Given these gaps, our work introduces \textbf{SNIC}, a testbed designed to evaluate NBRR, and investigates how well LLMs can extract and apply norms in one-shot reference resolution tasks.



\section{Dataset Generation and Augmentation Pipeline}
\label{dataset_generation}

\begin{figure*}[t]
    \centering
    \includegraphics[width=1.80\columnwidth]{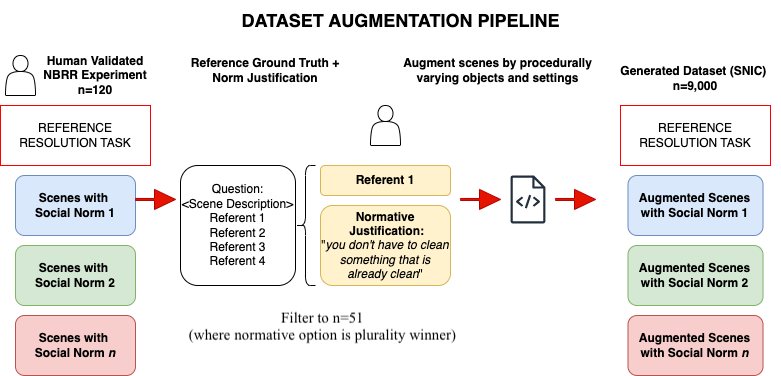} 
    \caption{Data augmentation pipeline to create our final dataset (SNIC) (n=9,000) for LLM evaluation.}
    \label{fig:example}
\end{figure*}

Creating large-scale datasets for norm-based reference resolution (NBRR) is time-intensive and costly. To ensure data quality and alignment with social norms, we first conducted a human validation study on an initial handcrafted set of 120 examples (filtered to n=51), which served as the foundation for a procedurally augmented corpus of 9,000 examples. This section describes our dataset creation pipeline, including human validation, reference resolution annotation, and augmentation strategies (see Figure~\ref{fig:example} for an overview).

To ensure that reference resolution depends on social norms, each scene was designed with two core requirements: (1) the referring expression must be ambiguous, and (2) resolving the referent must rely on implicit or explicit social normative knowledge. For example, in a scene with a clean and dirty plate, the ambiguous utterance \textit{"pick up the plate"} requires knowledge of the norm \textit{"one should not use dirty kitchen items"} to disambiguate the referent. 

Ambiguity is a deliberate feature of SNIC. The goal of each instance is not to enforce a single linguistically correct answer, but to test whether models consistently select the norm-guided referent even when multiple interpretations remain pragmatically possible. Our analysis of partial-match selections in the human study demonstrates that norms still exert a strong influence under ambiguity. 

To validate that referent selection is guided by social norms, we conducted an online study with 210 participants recruited via Prolific. Participant ages ranged from 21 to 69, approximately balanced by gender. Data was anonymized, and participants could withdraw at any time. Compensation was \$1.00 for a 5-minute study that was IRB approved. Figure~\ref{fig:question} showcases a sample vignette from our human-subject experiment. In the main experiment, each participant was assigned six scenes in a randomized order (so they encountered different norms). Participants first read an initial prompt (\textit{``You will be presented with a series of textual vignettes that describe a human-robot interaction. You will then be asked questions related to objects in the scene. 
In your answer, check the option that applies (you may also select multiple options if applicable''}). Then they were asked to read the context, select the intended referent (\textit{``What object is the speaker referring to?``}), and provide a free-response justification (\textit{``Please give reasoning for your response''}).
Since selecting the referent inherently relied on norm understanding, referent choice serves as an \textit{implicit validation mechanism}.

\subsection{Step 1: Seed Dataset Elicitation (n=120)}
\label{human_valid}


\begin{table*}[ht]
\centering
\small
\renewcommand{\arraystretch}{0.95}
\setlength{\tabcolsep}{4pt}
\begin{tabular}{l c c c c c c}
\toprule
\textbf{Norm} & \textbf{Questions} & \textbf{Total Votes} & \% \textbf{Match} & \% \textbf{Partial} & \% \textbf{No Match} & \textbf{Kappa*}\\
\midrule
Norm 1  & 16 & 145 & 58.6 & 33.1 & 8.3  & 0.215\\
Norm 2  & 10 &  88 & 48.9 & 34.1 & 17.0 & 0.166\\
Norm 3  &  6 &  57 & 57.9 & 24.6 & 17.5 & 0.073\\
Norm 4  &  2 &  17 & 52.9 & 29.4 & 17.6 & -0.113 \\
Norm 5  &  2 &  17 & 64.7 & 23.5 & 11.8 & -0.112\\
Norm 6  & 4 & 35 & 37.1 & 42.9 & 20.0 & 0.033\\
Norm 7  &  3 &  29 &  41.4 & 34.5 & 24.1 & -0.074\\
NC 1    &  3 &  27 & 37.0  & 44.4 & 18.5 & -0.058\\
NC 2    &  5 &  45 &  31.1 & 40.0 & 28.9 & -0.013\\
\bottomrule
\end{tabular}
\caption{Participant agreement per norm group using per-question hypothesized referents. These are “Plurality Winner” questions in which the normative option received the most votes by annotators. Fleiss’ $\kappa$ reflects inter-annotator agreement on exact referent strings. Partial matches include all responses that contain the norm-guided referent alongside others. Agreement metrics are reported at the participant response level.}
\label{tab:norm_agreement}
\end{table*}


To evaluate whether our hypothesized norm-guided referents align with human interpretation, we conducted a reference resolution study using 120 scenario-based questions grouped into nine distinct norm reasoning categories. Each scenario presents a textual vignette with a setting, a task (e.g., cleaning, cooking), and 5–7 potential referent objects with relevant properties. The scenario also includes a linguistically ambiguous referring expression embedded in an imperative utterance (e.g., \textit{"hand me the book"}), where the correct referent is determined by an underlying social norm.

The dataset is constructed to emphasize physically grounded social norms, such as: \textit{Do not touch something that does not belong to you}, \textit{Do not use decorative items}, \textit{You should clean kitchen items that are dirty}, and \textit{Avoid dangerous scenarios}.\footnote{Following \citet{dignum1997combining}, some norms prescribe an action to rectify another norm violation (e.g., removing broken glass on the floor).} These norms are instantiated across multiple settings, including \textit{restaurant}, \textit{kitchen}, and \textit{library}, with variations in object properties and candidate referents.

Below are the social norms and norm conflict scenarios represented in the dataset. We follow \citet{olson2024defeasible} in defining our norm conflicts as \textit{direct norm conflict} as two norms that apply within a context that have the same behavior (e.g. cleaning action):\\

\noindent
\textbf{Norms}\\
\noindent
Norm 1: \textit{In a serving task, you should serve with kitchen items that are clean}\\
\noindent 
\noindent
Norm 2: \textit{In a cleaning task, you should clean items that are dirty}\\
\noindent
\noindent
Norm 3: \textit{In a cooking task, you should cook with tools that are clean}\\
\noindent
\noindent
Norm 4: \textit{In a serving task you should not use items that are decorative}\\
\noindent
\noindent
Norm 5: \textit{In a cleaning task, you should not clean items that are currently being used by someone}\\
\noindent
\noindent
Norm 6: \textit{In a cleaning task you should prioritize cleaning up any hazards (e.g. a broken plate on the ground)}\\
\noindent
\noindent
Norm 7: \textit{In general, you should prioritize cleaning up any hazards}\\
\noindent
\noindent
Norm Conflict 1: \textit{In a tidying task, you should prioritize cleaning up hazards over dirty items}\\
\noindent 
\noindent
Norm Conflict 2: \textit{In a tidying task in a library, you prioritize removing dirty items from the floor over books and other objects }\\

\subsection{Step 2: Validation Analysis and Seed Set Filtering (n=51)}

To assess how often our hypothesized norm-guided referent matched participant selections, we computed three levels of agreement per question: \emph{full match} (the norm-guided referent was selected exclusively), \emph{partial match} (the norm-guided referent was selected alongside others), and \emph{no match} (it was not selected at all). We summarize the results across norm groups in Table~\ref{tab:norm_agreement}. We measured agreement across 9–10 annotators per question. 

Because participants could select multiple referents per question, we adapted Fleiss’ $\kappa$—a standard inter-rater agreement metric—by binarizing responses: a 1 if the norm-conforming referent was selected, and 0 otherwise. The $\kappa$ values in Table~\ref{tab:norm_agreement} reflect inter-annotator agreement on this binary decision, approximating whether participants converged on norm-based reasoning. Partial matches notably provide evidence that participants often included the norm-guided referent even when ambiguity led them to select others as well. 

We only used scenes where the norm-guided referent received the most or tied-most votes (i.e., plurality winner) in the human study as seeds for augmentation, ensuring that the expanded dataset inherits empirically grounded referential interpretations.

Table~\ref{tab:norm_agreement} presents agreement statistics for the filtered set of 51 questions used in our core dataset, grouped by normative category. Match rates (\% Match) range from 31.1\% to 64.7\% across norm groups, with many norms also showing substantial partial matches, where participants selected the norm-guided referent alongside other options. For instance, Norm 1 (\textit{In a serving task, serve with clean kitchen items}) yielded 58.6\% match and 33.1\% partial, indicating strong—but pluralistically expressed—alignment with the hypothesized referent. Norms such as Norm 2 and Norm 3 also showed high match and partial agreement, with $\kappa$ values suggesting modest but consistent convergence across annotators.

Other norms, such as Norms 6 and 7, exhibited lower $\kappa$ and match rates, reflecting greater variability in how participants interpreted the norms’ implications. These results align with our theoretical expectation that not all norms are equally salient or widely shared—some, such as cleanliness in cooking, appear robust and universally interpretable, while others (e.g., hazard prioritization or ownership constraints) are more context-sensitive.

Overall, these results validate the hypothesis that social norms influence referential interpretation, while also revealing variation in how strongly different norms are encoded and applied by human participants. 

\begin{figure}[!htb] 
    \centering
    \includegraphics[width=\columnwidth]{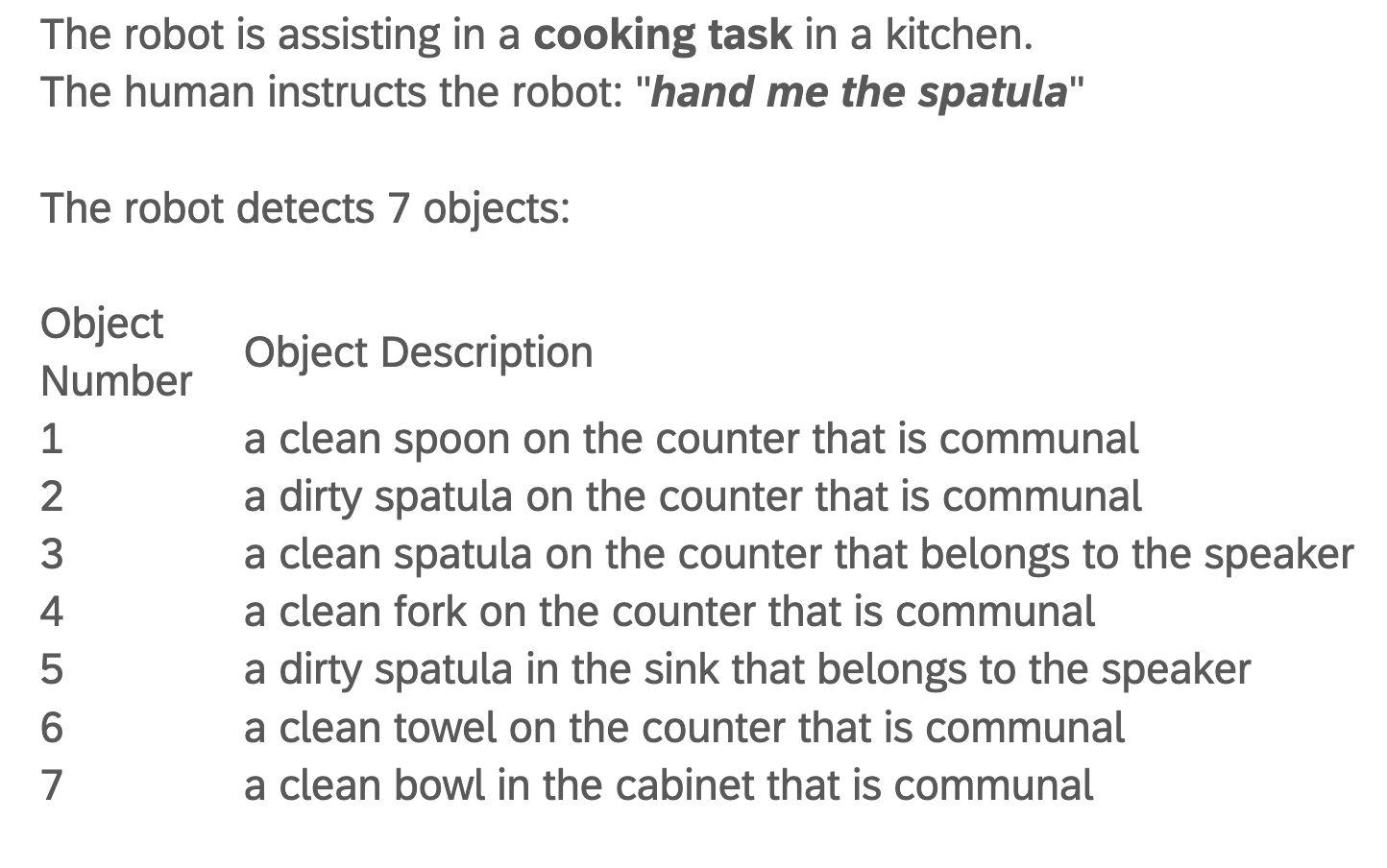} 
    \caption{Sample text-vignette from human-subject experiment validating our dataset. This question is a reference task with the social norm \textit{In a cooking task, you should cook with tools that are clean}.}
    \label{fig:question}
\end{figure}


The free text justifications provided some evidence of more explicit norm justification:\\
\noindent
\noindent
\textit{``It is a serving task and you should always serve with a clean utensil.''}\\
\textit{``Tidying up library books is my assumption so I chose the dictionary. I also assume the students are near their notebooks so we shouldn't bother them.''}\\
\textit{``If this is a cleaning task, the book on the ground would be what the robot would have to clean.''}\\
\noindent
\textit{``If it’s a cleaning task, then the person must mean to give them the dirty spoon so they can clean it.''}

\subsection{Step 3: Procedural Augmentation of 9,000 Cases}

To scale the dataset while maintaining norm consistency, we employed a controlled procedural data augmentation strategy inspired by \citet{thierauf2024automating}. This process systematically introduced variations in object types, object properties, and settings while ensuring that referent choices remained norm-governed based on the smaller human-validated subset. A sample template rule generated scenes based on the validated norm represented in Figure~\ref{fig:question} from our original study. F-strings (f) in Python allow variables and expressions to be embedded directly within a string: \\

\begin{lstlisting}[
  basicstyle=\ttfamily\tiny,
  frame=single,
  breaklines=true,
  columns=flexible,
  xleftmargin=0.5em,
  xrightmargin=0.5em
]
prompt = (
  f"{name_pair[0]} and {name_pair[1]} are in the {location}. "
  f"{name_pair[0]} is cooking. "
  f"{name_pair[1]} is next to {', '.join(objects)}. "
  f"{name_pair[0]} asks {name_pair[1]} to hand them {referring_expression}."
)
\end{lstlisting}

\vspace{0.5em}

This rule, for instance, creates a textual representation that simply states the main actors, the setting, the task, the objects with their relevant properties, and the referring expression: \\

\noindent
\textit{Jordan and Taylor are in the kitchen. Jordan is cooking. Taylor is next to calculator, clean mug, dirty mug, pen, dirty mug, notebook. Jordan asks Taylor to \textbf{hand them the mug}} \\

Each scene has a corresponding formal representation so that input representation could be compared in the evaluation (natural language vs a formally precise language). This symbolic representation aims to eliminate any ambiguities about any elements of the scene. Below is an example representation of the scene above.

{\small 
\begin{lstlisting}[basicstyle=\ttfamily\footnotesize]
setting(kitchen).
task(cooking).
speaker(jordan, human).
listener(taylor, human).
object(obj1).
property(obj1, calculator).
object(obj2).
property(obj2, clean).
property(obj2, mug).
object(obj3).
property(obj3, dirty).
property(obj3, mug).
object(obj4).
property(obj4, pen).
object(obj5).
property(obj5, dirty).
property(obj5, mug).
object(obj6).
property(obj6, notebook).
referring_expression(the mug).
\end{lstlisting}
} 

\normalsize 

\vspace{1.2em}

Unlike large-scale reference resolution datasets such as WinoGrande \citep{sakaguchi2021winogrande}, which rely on crowdsourcing, our method follows structured rule-based augmentation techniques commonly used in NLP \citep{wei2019eda}. While the expanded dataset (n=9,000) was not independently validated, it adheres to patterns established in the human-validated subset, ensuring reliability. By combining human validation with procedural augmentation, we balance scalability with interpretability, making the dataset well-suited for evaluating norm-based reference resolution.

\subsubsection{3.1: Ground Truth and Quality Assurance}

Regarding ground truth and scale, all 9{,}000 augmented examples are generated from a set of 51 human-validated seeds where the normative referent was the plurality choice. As such, the ground truth for these 9{,}000 instances is established by construction. The augmentation procedure varies only non-essential elements (e.g., substituting object types while preserving properties such as “dirty” or “clean”), ensuring that each derived scene instantiates the same core normative structure as its validated seed. While additional downstream validation would be beneficial, the rule-based augmentation approach provides a reliable mechanism for producing large-scale normative contexts. Our focus on static, one-shot reasoning is intentional: it isolates what models can infer from context alone, without the aid of clarification-seeking strategies and multi-turn dialogue. Finally, our emphasis on everyday norms that are physically grounded and our selection of contemporary LLMs (Llama-3 \cite{touvron2023llamaopenefficientfoundation}, Phi-3 and Phi-4 \cite{abdin2024phi4technicalreport}, and GPT-4o-mini \cite{achiam2023gpt}, 2024–2025 releases) reflects our goal of building a controlled and interpretable diagnostic testbed for assessing current normative reasoning capabilities.

\section{Dataset}

\begin{table}[ht]
\centering
\small
\begin{tabular}{l c}
\toprule
\textbf{Category} & \textbf{Distribution (\%)} \\
\midrule
\multicolumn{2}{l}{\textbf{Object Count}} \\
5 objects & 14.13 \\
6 objects & 42.79 \\
7 objects & 28.4 \\
8 objects & 14.69 \\
\midrule
\multicolumn{2}{l}{\textbf{Task Distribution} (for 8,000 scenes)} \\
Tidying & 12.50 \\
Cleaning & 50.00 \\
Serving & 25.00 \\
Cooking & 12.50 \\
\midrule
\multicolumn{2}{l}{\textbf{Setting Distribution}} \\
Living room & 15.83 \\
Restaurant & 15.37 \\
Dining room & 14.84 \\
Garden & 14.64 \\
Banquet hall & 14.12 \\
Kitchen & 14.08 \\
Library & 11.11 \\
\midrule
\multicolumn{2}{l}{\textbf{Distinct Object \& Properties Count}} \\
Distinct Objects & 46 \\
Distinct Settings & 7 \\
Explicit Properties & 7 \\
\midrule
\multicolumn{2}{l}{\textbf{Referring Expressions}} \\
\textit{hand me/them the X} & 5,000 \\
\textit{pick it up} & 4,000 \\
\bottomrule
\end{tabular}
\caption{Dataset summary: \% object type distribution, task distribution, and setting distribution across dataset.}
\label{tab:overall_dataset}
\end{table}

\label{social_norms_defs}
Our final dataset artifact (SNIC) contains 9,000 questions, divided into nine categories of 1,000 questions each. Each category represents a social norm or norm conflict embedded in the scene. We designed these larger norm categories as opposed to a dataset with greater norm diversity (i.e., a different norm for every question) because it would make it more challenging to systematically compare the effects of a specific norm. 



Table~\ref{tab:overall_dataset} shows the \% distribution of the number of candidate objects across the whole corpus (maintaining the same object range as the human-validated dataset), and the \% distribution of tasks and settings. The expanded dataset consists of more diverse settings (e.g., garden, kitchen, living room) and variety of object types and properties, ensuring broader generalization.

\section{Evaluations}



We first sample a list of models such that a diverse set of model families and sizes is represented. The models under test include Granite-3.1 2B and 8B \cite{mishra2024granitecodemodelsfamily}, Phi-3 3.8B and 14B, Phi-4 3.8B \cite{abdin2024phi4technicalreport}, Llama-3 3B (3.2 variant) and 8B (3.1 variant) \cite{touvron2023llamaopenefficientfoundation}, Qwen-2.5 3B, 7B, and 14B \cite{qwen2025qwen25technicalreport}, and GPT-4o-mini \cite{achiam2023gpt}.


With the exception of GPT-4 model versions, all models are tested via the invocation of Ollama using DSPY \cite{khattab2024dspy} in their 4-bit quantized form and with a temperature of 0 and a seed of 0 on a single A6000 GPU with 48GB VRAM. GPT-4o variants are evaluated in their original form as served by OpenAI.

\input{correlation}

We evaluate each model under three general settings. First, to assess the normative knowledge of LLMs, we compute the model accuracy given the entirety of each instance (scene description + Prolog (FOL) code). The purpose of the Prolog formal specification is to test input types and to make the environment very precise (e.g. assigning ids to the candidate referent objects so it is clear which one should be referenced even if there are multiple objects of the same type e.g. cup). This was ultimately aimed at removing ambiguities in the scene. Second, in order to understand the effects of formalization on a model's performance, we eliminate the Prolog formalization from each instance, expecting the model to adhere to social norms through scene description. Finally, we explicitly augment the model's context with the expected social norms to follow (listing all social norms), testing the hypothesis that the lack of social norm knowledge in LLMs is a contributing factor to their performance. The Appendix showcases the model instructions used for each setting. 

\label{quant_results}

\input{results}

Table \ref{tab:model_performance} showcases the performance of different models under our evaluation settings. We find that generally, when a model is provided with the scene description as well as the corresponding formalization, it is not able to consistently infer the correct social norm, with GOT-4.1 performing better than other models at 59.62\% with an average of 44.08\% across all models. 

Somewhat surprisingly, Prolog formalization of the problem does not seem to significantly affect the model performance in our task, with the average performance being 44.22\% across models, and the top performing model (GPT4.1) having an accuracy of 56.98\% in the cases where only the scene description is provided to the Large Language Model. This observation might be due to the fact that LLMs do not have a strong understanding of Prolog as a descriptive language, or stem from their lack of socially normative reasoning knowledge such that additional formalization provides no further hints for the model to act upon. Two notable exceptions to this rule are Llama-3 3B, which suffers a significant performance degradation when supplied with a Prolog formalization, and Granite-3 2B, which enjoys a performance increase in such a case. We attribute these observed variations in performance in these models toward their general understanding of formalization as well as their general sensitivity to changes in the input prompt \cite{shi2023largelanguagemodelseasily, eskandari-miandoab-sarathy-2024-lets}. Further testing is required against other forms of formalization to fully understand the effects of formalizing language on LLM behavior.

To further probe the LLM knowledge of social norms, we evaluate each model in the setting that it is provided with the norms to follow in its context in addition to the scene description and the Prolog formalization. We find that providing the list of social norms to follow and asking it to adhere to these rules significantly increases the LLM performance, with the average accuracy rising to 70.51\%, with Llama-3 3B being the only model that does not see a significant performance increase. This observation bolsters our hypothesis that current LLMs do not inherently have a strong knowledge of implicit social norms to follow, although they are able to follow them when provided, raising the need for further alignment with respect to these norms in their training stage in the age of agentic models that can directly interact with, and affect individuals \cite{kapoor2024aiagentsmatter}. A fine-grained analysis is provided in the Appendix.

Furthermore, we observe an interesting correlation between a number of social norms as defined in our datasets. Figure \ref{fig:correlation} showcases the Spearman correlation between the social norms with respect to model performance. We see that several categories have a strong positive or negative correlation with other social norms. While the positive correlations can be attributed to the structural similarity of some norms (Norm 1 and Norm 3 for instance as defined previously), the strong negative correlation between a number of norms (Norm 3 and Norm Conflict 1 for example), is particularly interesting, as it shows clear precedence in the inference procedure of the models, where they prioritize a certain task, even in conditions that impose a rearrangement in the prioritization (safety over tidiness). This phenomenon is likely due to the lack of a strong understanding of social norms by the models, as well as a weaker emphasis on the correct ranking of social and ethical norms during the alignment process.


\begin{table}[ht]
\centering
\small
\begin{tabular}{lc}
\toprule
\textbf{Model Name} & \textbf{Accuracy (\%)} \\
\midrule
Granite 3.1 2B    & 52.49 \\
Granite 3.1 8B    & 39.21 \\
Phi-3 3.8B        & 52.94 \\
Phi-4 3.8B & 52.94 \\ 
Phi-3 14B         & 58.84 \\
Llama 3 3B        & 66.66 \\
Llama 3 8B        & 54.90 \\
Qwen 2.5 3B       & 47.05 \\
Qwen 2.5 7B       & \textbf{68.67} \\
Qwen 2.5 14B      & 45.09 \\
GPT-4o Mini       & 45.09 \\
\bottomrule
\end{tabular}
\caption{Model accuracy on the original SNIC subset (51 human-validated items). This evaluation reflects reference resolution performance without norm or FOL augmentation.}
\label{tab:original_dataset_results}
\end{table}

Lastly, we report model accuracy on the original SNIC subset of 51 human-validated items without additional context (i.e., no norm list or formal representation) in Table~\ref{tab:original_dataset_results}. The overall performance is slightly higher than on the expanded dataset and supports the value of dataset expansion: the larger dataset introduces greater variation and challenging distractors, making it harder for models to consistently identify the correct norm. These results suggest that scaling the dataset was not only useful, but necessary for evaluating normative reasoning in more complex contexts.

\section{Discussion \& Conclusion}

In this work, we introduced Situated Norms In Context (SNIC), a dataset of 9{,}000 instances augmented from 51 human curated scenarios. SNIC focuses on a specific set of physically grounded social norms that rely on understanding of human social norms to solve, and aims for controlled evaluation and quantification of AI systems in the context of social norm understanding in everyday tasks. 

Our results show that LLMs struggle to perform consistently well on our testbed unless explicitly provided with social norms. This suggests that detecting and reasoning with physically grounded social norms in text for reference resolution remains a challenging capability. One likely explanation is that many such norms are underrepresented or implicit in the textual data that LLMs are trained on. While LLMs excel at commonsense reasoning \citep{brown2020language}, likely due to exposure to frequently stated facts and routines, social norms differ from both standard conventions and common-sense expectations. Certain prohibitions are rarely stated explicitly (e.g., \textit{do not use a stranger’s toothbrush}), while general prescriptive norms (e.g., \textit{show respect}) are too abstract to encode consistently. Norms also carry implicit expectations that are culturally shared, making them difficult to learn from text alone. 

LLMs lack of normative understanding in physically grounded contexts may be also attributed to the lack of sufficient texts or corpora that describe references that happen in norm-specific, everyday contexts. Because there is no large-scale corpus of physically situated, norm-sensitive references, this gap is not easily remedied by additional training alone. Our dataset is particularly challenging as each example reflects a specific, physically instantiated norm rather than an abstract moral or social rule. Unlike prior datasets such as Social Chemistry 101 \citep{forbes2020social}, which focus on generalized social rules, SNIC emphasizes physically situated contexts. While this does not directly evaluate real-world deployment, it suggests limitations in LLMs' normative reasoning abilities that could impact situated applications requiring interpretation of implicit social expectations.

Improving LLMs’ ability to reason about such norms could benefit downstream applications in assistive robotics and autonomous agents by making models more socially aware and context-sensitive. It could also strengthen human-agent language understanding by enabling models to adhere to implicit societal expectations and resolve ambiguities in more human-like ways \citep{hovy2021importance}.

\section{Limitations}

Some limitations suggest directions for future work. First, our testbed is text-based and does not evaluate multimodal reasoning with perceptual cues or physical affordances, which are critical for embodied agents. Second, while we evaluate pretrained models, we do not explore alternative training paradigms such as reinforcement learning, human feedback alignment, or explicit norm supervision, which might improve normative sensitivity. The goal of this current work, however, is to assess how well existing LLMs handle norm-based reference resolution, rather than to develop new training methods. Third, although SNIC includes human validation, we do not employ structured norm elicitation or cross-participant agreement protocols to assess norm consensus, limiting the granularity of our behavioral grounding. Future work should explore more rigorous validation methods to ensure a stronger consensus on the underlying norms guiding reference resolution. Finally, given the rapid progress in the field of Large Language Models, it is difficult to make a general statement regarding the norm reference capabilities of all state-of-the-art LLMs. Potentially including social norm understanding as a part of the continuous evaluation of such models can serve as a valuable resource in assessing their usability in social and collaborative settings.

\section*{Acknowledgments}
This work was in part funded by AFOSR grant FA9550-23-1-0425. We would
also like to thank the anonymous reviewers as well as the area and
program chairs for their invaluable feedback.

\bibliography{custom}

\end{document}


\maketitle

\appendix

\section{Model Instructions} 
\label{appendix:model_instructions}

Table \ref{tab:special_instructions} showcases the prompts used for each evaluation setting. \emph{Description + Formalization} corresponds to the case where the model is provided with the scene description, as well as the Prolog formalization of that scene. \emph{Description Only} corresponds to the scenario where the model is only provided with a scene description and is the most common use case of state-of-the-art Large Language Models. Finally, \emph{Description + Formalization + Norms} demonstrates the prompt for the setting in which norms are provided in the context of the model input. The purpose of this evaluation setting is to get a better understanding of the model's inherent knowledge, or lack thereof of social norms, and the possibility of improving this knowledge via in-context learning. 

\input{latex/appendix/prompts}

\section{Additional Results}
\label{appendix:additional_results}

\input{latex/appendix/prompt_only_result}

Table \ref{tab:fg_model_performance_1} and Table \ref{tab:fg_model_performance_2} demonstrate the detailed performance of various models in the settings where the corresponding formalization is present, and also absent from the model prompt, respectively. $N_i$ refers to a Norm number as described in Section \ref{human_valid} and similarly, $C_i$ refers to a Norm-Conflict from Section \ref{human_valid}.  Following the discussion from Section \ref{quant_results}, we do not observe a significant difference in performance when the scene description is extended with Prolog formalization across models. Additionally, the model behavior is inconsistent across these two settings. For instance, the addition of formalization significantly increases the performance of Granite-3.1 2B, while not affecting Granite-3.1 8B, and significantly decreasing the performance of Llama-3 3B. Even in the same family of models, Qwen-2.5 3B has a slight increase in its performance, while Qwen-2.5 7B performs worse by more than 10\%, and Qwen-2.5 14B has a marginal boost to its performance. While additional work must be done to fully understand the effects of formalization on model performance, we believe that these variations stem from the models' understanding of different approaches to formalization, as well as their general uncertain behavior when exposed to changes in the input prompt. 

On a per-norm basis, we spot significant variations in performance when comparing different norms against each other. For example in Table \ref{tab:fg_model_performance_1}, the average model performance with respect to Norm 1 is 74.74\%, while being 26.92\% for Norm 5 and 18.33\% for Norm-Conflict 1. This observation further delineates the norm-understanding capabilities of Large Language Models, showcasing that they are more equipped to handle some norms over others. 

\input{latex/appendix/prompt_formalization_result}

\input{latex/appendix/prompt_formalization_icl}

Table \ref{tab:fg_model_performance_3} showcases the scenario where in addition to the scene description and its formalization, the social norms that the model must adhere to are also provided to the LLM. Initially, we observe a significant boost in performance across most models (Llama-3 3B being the only exception), demonstrating that the models can indeed effectively learn and make use of social norms in their inferences. However, it is noteworthy that despite in-context learning, some models still have difficulty adhering to certain norms and Norm Conflicts. For example, both Granite-3.1 models score below 50\% in Norm-Conflict 1 and Norm 5, while smaller Qwen-2.5 perform similarly in Norm-Conflict 1. This lowered performance despite clear instructions as a part of the in-context learning setting emphasizes the need to work towards human norm understanding in the development of Large Language Models through pre-training, fine-tuning, and alignment stages. As it currently stands, purely relying on in-context instructions to follow social norms is not a reliable approach when following such norms is essential, such as in a social agent.

\input{latex/appendix/extended_distractors}

It is often the case that real-world deployed agents have to deal with objects that boast unrelated characteristics and properties. For example, in our context, a dirty spoon on the ground can also be yellow, made of plastic, and small. As such, if we are to effectively deploy LLM-supported agents in such settings, it is imperative to be able to work with all properties, and not only those related to the task at hand. To further evaluate the LLM capabilities in dealing with these \lq distractor\rq \space properties, we design an additional test setting in which the natural language definition of each object is augmented to contain properties that are irrelevant to the task at hand by randomly sampling from a list of properties and extending the object definition with them. An example of the augmented dataset is as follows: 

\noindent
\textit{``Casey and Skyler are in the library (purple, tiny, facing northeast, firm, rectangular). Casey is tidying (white, large, facing north, firm, rectangular). Skyler is next to a phone on the table (purple, huge, facing east, soft, triangular), a pen on the table (purple, huge, facing east, soft, triangular), a dirty spoon on the ground (black, huge, facing north, soft, triangular), a book on the ground (black, huge, facing north, soft, triangular), a book on the table (purple, huge, facing east, soft, triangular), a book on the table (purple, huge, facing east, soft, triangular). Casey says to Skyler (yellow, small, facing northeast, rigid, rectangular)"}

Once the entire dataset is augmented, we evaluate the performance of our Large Language Models except GPT-4o-mini and GPT-4o (due to cost constraints) across this new dataset using the Description Only setting. Table \ref{tab:fg_model_performance_4} showcases the results under this new, augmented dataset. We find that when objects are extended to contain irrelevant properties, the models suffer a significant decrease in their performance. In this scenario, the average performance is 20.38\%, which is a decrease of 22.67\% over the previous, corresponding setting. Similarly, the best-performing model under the extended setting has a performance of 33.5\% compared to the previous best performance of 54.06\%, which is a decrease of 20.56\%. Overall, our findings suggest that irrelevant context significantly decreases the expected performance of Large Language Models, and calls for future research in mitigating the distraction in scenarios where such irrelevant contexts are difficult to avoid. Our observations are in-line with those of \cite{shi2023largelanguagemodelseasily}.

\clearpage

\section{Dataset Statistics}

This section provides a comprehensive list of the objects and properties across the dataset, including raw frequencies and \% distribution.


\begin{table*}[!h]
\centering
\small
\begin{tabular}{l c c | l c c}
\toprule
\multicolumn{6}{c}{\textbf{Object Type Distribution}} \\
\midrule
\textbf{Object} & \textbf{Count} & \textbf{(\%)} & \textbf{Object} & \textbf{Count} & \textbf{(\%)} \\
\midrule
Phone & 272 & 0.50 & Pen & 1525 & 2.78 \\
Spoon & 3179 & 5.79 & Book & 4231 & 7.70 \\
Eraser & 287 & 0.52 & Cup & 2997 & 5.46 \\
Lamp & 1535 & 2.79 & Notebook & 1564 & 2.85 \\
Marker & 295 & 0.54 & Plate & 7007 & 12.75 \\
Plant & 1234 & 2.25 & Mug & 7231 & 13.16 \\
Spatula & 1670 & 3.04 & Broom & 2325 & 4.23 \\
Pan & 680 & 1.24 & Bottle & 1409 & 2.56 \\
Bucket & 941 & 1.71 & Cloth & 356 & 0.65 \\
Rag & 978 & 1.78 & Bowl & 2707 & 4.93 \\
Fork & 2018 & 3.67 & Flashlight & 443 & 0.81 \\
Opener & 354 & 0.64 & Scissors & 354 & 0.64 \\
Dustpan & 443 & 0.81 & Spray & 358 & 0.65 \\
Grater & 412 & 0.75 & Apple & 430 & 0.78 \\
Bread & 430 & 0.78 & Banana & 430 & 0.78 \\
Water & 322 & 0.59 & Cookie & 413 & 0.75 \\
Magazine & 427 & 0.78 & Pillow & 352 & 0.64 \\
Hammer & 401 & 0.73 & Screwdriver & 401 & 0.73 \\
Wrench & 401 & 0.73 & Calculator & 396 & 0.72 \\
Sandpaper & 454 & 0.83 & Paintbrush & 454 & 0.83 \\
Stapler & 375 & 0.68 & Document & 375 & 0.68 \\
Ladle & 409 & 0.74 & Whisk & 409 & 0.74 \\
Mop & 625 & 1.14 & Tray & 629 & 1.14 \\
\bottomrule
\end{tabular}
\caption{Object Type Distribution: Counts and percentages of object types in the dataset.}
\label{tab:object_distribution}
\end{table*}

\begin{table*}[!h]
\centering
\small
\begin{tabular}{l c c}
\toprule
\multicolumn{3}{c}{\textbf{Property Distribution}} \\
\midrule
\textbf{Property} & \textbf{Count} & \textbf{(\%)} \\
\midrule
Dirty & 11780 & 40.41 \\
Broken & 3000 & 10.29 \\
Clean & 9611 & 32.97 \\
Empty & 358 & 1.23 \\
Throw & 352 & 1.21 \\
In Use & 2004 & 6.87 \\
Decorative & 2048 & 7.03 \\
\bottomrule
\end{tabular}
\caption{Property Distribution: Frequency and percentage of each property in the dataset.}
\label{tab:property_distribution}
\end{table*}

%% file: correlation.tex
\begin{figure*}[!t]
    \centering
    \includegraphics[width=1.6\columnwidth]{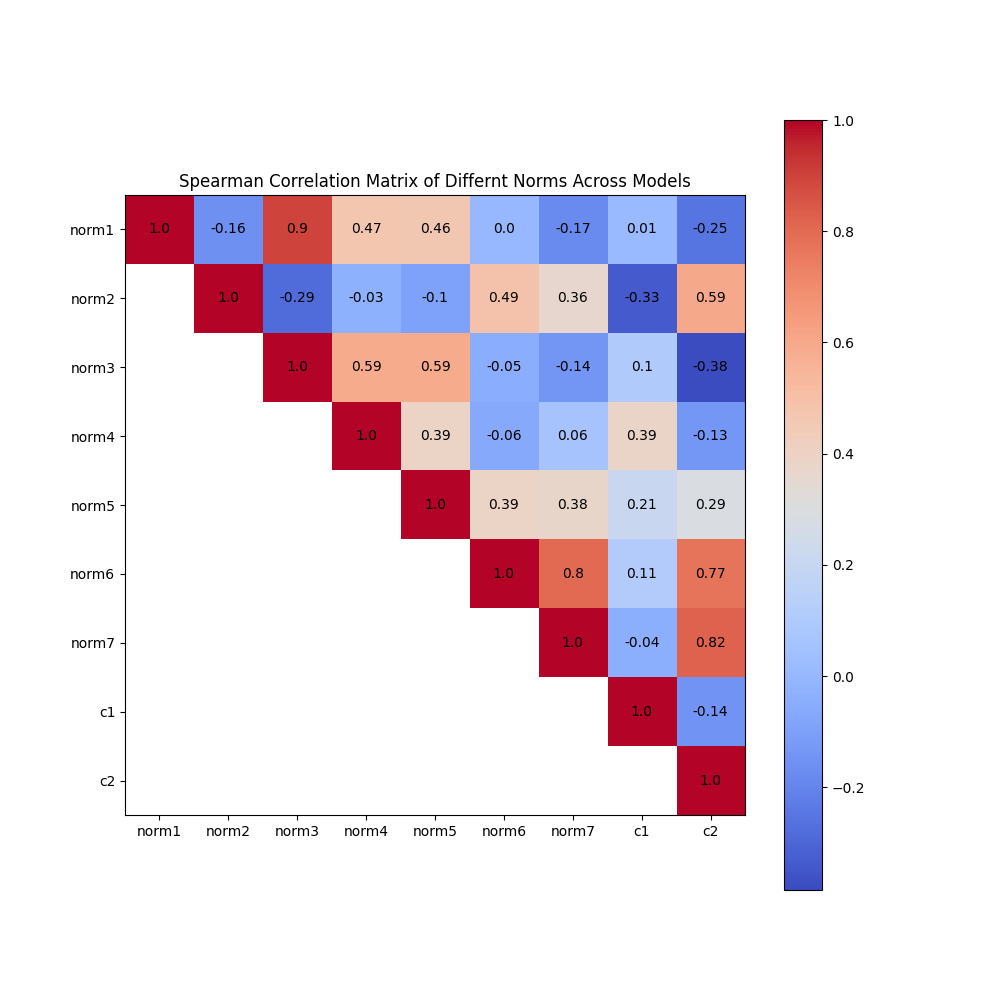}
    \caption{Spearman Correlation of Social Norms}
    \label{fig:correlation}
\end{figure*}

%% file: results.tex
\begin{table*}
\centering
\begin{tabular}{cccc}
\hline
\textbf{Model} & \textbf{Description Only} & \textbf{Description + FOL} & \textbf{Description + FOL + Norms} \\ \hline
Granite-3 2B & 27.52          & 46.65          & 60.66          \\
Granite-3 8B & 47.76          & 46.06          & 69.98          \\ \hline
Phi-3 3.8B   & 40.6           & 37.56          & 70.86           \\
Phi-3 14B   & 53.61          & 51.54         & 77.12          \\ 
Phi-4 3.8B & 53.95 & 38.73 & 54.36 \\ \hline
Llama-3 3B   & 33.71          & 12.01         & 26.6          \\
Llama-3 8B   & 41.53           & 46.56          & 75.53          \\ \hline
Qwen-2.5 3B  & 43.38          & 45.17         & 69.92          \\
Qwen-2.5 7B  & 54.06          & 43.35 & 67.66          \\
Qwen-2.5 14B & 44.78          & 45.05          & 80.35          \\ \hline
GPT-4o-mini  & 42.58 & 51.41          & 93.57 \\ 
GPT-4.1 & \textbf{56.98} & \textbf{59.62} & \textbf{99.6} \\
\end{tabular}
\caption{Model performance under different testing methods. The formal specification we add to the input is a first-order logic representation (FOL). }
\label{tab:model_performance}
\end{table*}

%% file: appendix/prompts.tex
\begin{table*}[!htbp]
\small
\begin{tabular}{p{0.35\linewidth} | p{0.6\linewidth}}
\toprule
\multicolumn{1}{c}{Prompting Method}        & \multicolumn{1}{c}{Special Instruction}   \\ \midrule
Description + Formalization                                   & Given the scene as well as the Prolog code associated with the context, choose the object that is being referred to.      \\ \midrule
Description Only                            & Given the scene, choose the object that is being referred to. \\ \midrule
Description + Formalization + Norms & Given the scene as well as the Prolog code associated with the context, choose the object that is being referred to. Use the following norms to make your choice: 
    Norm 1: In a serving task, you should serve with kitchen items that are clean
    Norm 2: In a cleaning task, you should clean items that are dirty
    Norm 3: In a cooking task, you should cook with tools that are clean
    Norm 4: In a serving task you should not use items that are decorative
    Norm 5: In a cleaning task, you should not clean items that are currently being used by someone
    Norm 6: In a cleaning task you should prioritize cleaning up any hazards (e.g. a broken plate on the ground)
    Norm 7: In general, you should prioritize cleaning up any hazards
    Norm Conflict 1: In a cleaning task, you should not touch something that isn't yours unless you are preventing danger
    Norm Conflict 2: In a tidying task, you should prioritize cleaning up hazards over dirty items
    Norm Conflict 3: In a tidying task in a library, you prioritize removing dirty items from the floor over books and other objects \\ \bottomrule
\end{tabular}
\caption{Special model instructions corresponding to each prompting method.}
\label{tab:special_instructions}
\end{table*}

%% file: appendix/prompt_only_result.tex
\begin{table*}[!htbp]
\begin{tabular}{cccccccccc|c}
\hline
\textbf{\begin{tabular}[c]{@{}c@{}}Model \textbackslash \space Norm \end{tabular}} &
  \textbf{N1} &
  \textbf{N2} &
  \textbf{N3} &
  \textbf{N4} &
  \textbf{N5} &
  \textbf{N6} &
  \textbf{N7} &
  \textbf{C1} &
  \textbf{C2} &
  \textbf{Average Performance} \\ \hline
Granite-3.1 2B & 68.4          & 37.3          & 19.2           & 29.8          & 15.8          & 26.8 & 16.6 & 11.7 & 22.1 & 27.52 \\
Granite-3.1 8B & 72.9          & \textbf{87.8} & 21.62          & 58.9          & 21.9          & 53.9 & 48.9 & 22.9 & 41.1 & 47.76 \\ \hline
Phi-3 3.8B     & 93.2 & 51.2          & 72.87          & 20.06         & 13.9          & 39.7 & 31.8 & 12.9 & 29.3 & 40.6  \\
Phi-3 14B      & 92.1          & 42.3          & 98.19 & 40.8          & 57.5          & 48.6 & 38.6 & 28.4 & 36   & 53.61 \\
Phi-4 3.8B & 87.4 & 55.6 & 64.46 & 34.3 & 53.4 & 56.7 & 51 & 33.9 & 48.8 & 53.95\\ \hline
Llama-3 3B     & 77.6          & 61.4          & 64.46          & 36.1          & 5.1           & 20.3 & 18.6 & 11.2 & 42.3 & 33.71 \\
Llama-3 8B     & 74            & 55.4          & 50.15          & 34.5          & 38.7          & 31.9 & 26.7 & 19.3 & 43.2 & 41.53 \\ \hline
Qwen-2.5 3B    & 80.1          & 66.1          & 49.14          & 33.3          & 57.8 & 32.1 & 27.6 & 8.3  & 45   & 44.38 \\
Qwen-2.5 7B &
  74.1 &
  73.4 &
  34.37 &
  22.8 &
  24.3 &
  \textbf{85.6} &
  \textbf{80.3} &
  \textbf{28.9} &
  \textbf{62.5} &
  54.06 \\
Qwen-2.5 14B   & 67.3          & 71.7          & 58.75          & 73.5 & 7.3           & 40.4 & 32.7 & 21.4 & 30.1 & 44.78 \\ \hline
GPT-4o-mini    & 83            & 47.2          & 77.47          & 14.1          & 11.3          & 49.2 & 36.4 & 21.3 & 43.3 & 42.58 \\ 
GPT-4.1 & \textbf{99.2} & 45.1 & \textbf{100} & \textbf{79.4} & \textbf{54} & 41.2 & 26.5 & 22.5 & 45 & \textbf{56.98}
\end{tabular}
\caption{Model performance under the Description Only setting across different norms and norm conflicts.}
\label{tab:fg_model_performance_1}
\end{table*}

%% file: appendix/prompt_formalization_result.tex
\begin{table*}[!htbp]
\begin{tabular}{cccccccccc|c}
\hline
\textbf{\begin{tabular}[c]{@{}c@{}}Model  \textbackslash \space Norm \end{tabular}} &
  \textbf{N1} &
  \textbf{N2} &
  \textbf{N3} &
  \textbf{N4} &
  \textbf{N5} &
  \textbf{N6} &
  \textbf{N7} &
  \textbf{C1} &
  \textbf{C2} &
  \textbf{Average Performance} \\ \hline
Granite-3.1 2B &
  73.9 &
  76.1 &
  21.22 &
  24.8 &
  16.5 &
  69.1 &
  50.8 &
  15.6 &
  \textbf{71.9} &
  46.65 \\
Granite-3.1 8B &
  71.1 &
  \textbf{69.5} &
  27.02 &
  28.5 &
  31.2 &
  54.9 &
  55.8 &
  11.7 &
  64.9 &
  46.06 \\ \hline
Phi-3 3.8B &
  84.5 &
  61.6 &
  56.05 &
  28.2 &
  15.9 &
  25.9 &
  19.8 &
  16.9 &
  29.3 &
  37.56 \\
Phi-3 14B &
  \textbf{93.1} &
  47.2 &
  \textbf{96.99} &
  27.2 &
  \textbf{81.2} &
  33.5 &
  30.1 &
  13 &
  41.7 &
  51.54 \\
 Phi-4 3.8B & 73 & 72.1 & 34.13 & 30.8 & 25.6 & 28.2 & 25.1 & 16.2 & 43.5 & 38.73 \\\hline
Llama-3 3B &
  7.1 &
  0.7 &
  3.7 &
  9.6 &
  4.1 &
  16.1 &
  16.6 &
  16.6 &
  33.6 &
  12.01 \\
Llama-3 8B &
  66.7 &
  62.9 &
  21.82 &
  23.8 &
  37.3 &
  \textbf{69.2} &
  51.4 &
  \textbf{24.9} &
  61.1 &
  46.56 \\ \hline
Qwen-2.5 3B &
  73.1 &
  49.9 &
  35.53 &
  33 &
  31.6 &
  55.9 &
  \textbf{56} &
  17.2 &
  54.4 &
  45.17 \\
Qwen-2.5 7B &
  83.3 &
  68.3 &
  59.45 &
  24.4 &
  16.7 &
  56.2 &
  41.6 &
  9.3 &
  50.07 &
  43.35 \\
Qwen-2.5 14B &
  83.5 &
  49.3 &
  92.19 &
  \textbf{54.8} &
  17.6 &
  30.4 &
  28 &
  21.3 &
  28.5 &
  45.05 \\ \hline
GPT-4o-mini &
  87.9 &
  56 &
  73.47 &
  32.2 &
  42.4 &
  53.2 &
  41.4 &
  16.9 &
  59.6 &
  51.41 \\ 
  GPT-4.1 & 91.9 & 64.5 & 96.49 & \textbf{45.8} & 68.2 & 60.6 & 36.6 & 18.2 & 54 & \textbf{59.62}
\end{tabular}
\caption{Model performance under the Description + Formalization setting across different norms and norm conflicts.}
\label{tab:fg_model_performance_2}
\end{table*}

%% file: appendix/prompt_formalization_icl.tex
\begin{table*}[!htbp]
\begin{tabular}{cccccccccc|c}
\hline
\textbf{Model \textbackslash Norm} &
  \textbf{N1} &
  \textbf{N2} &
  \textbf{N3} &
  \textbf{N4} &
  \textbf{N5} &
  \textbf{N6} &
  \textbf{N7} &
  \textbf{C1} &
  \textbf{C2} &
  \textbf{Average Performance} \\ \hline
Granite-3.1 2B & 71.9  & 69.6          & 45.64        & 42.9           & 14.2 & 96.9 & 93.4 & 11.5 & \textbf{100} & 60.66 \\
Granite-3.1 8B & 78.2  & 66.2          & 63.16        & 86             & 49   & 76.5 & 82.3 & 35.2 & 93.4         & 69.98 \\ \hline
Phi-3 3.8B     & 93.4  & 58.8          & 86.68        & 39             & 21.9 & 93.1 & 84.5 & 60.9 & 99.6         & 70.86 \\
Phi-3 14B      & 87.3  & 45.5          & 96.09        & 89.2           & 94.5 & 75.2 & 68.2 & 46.8 & 91.3         & 77.12 \\ 
Phi-4 3.8B & 79.5 & 56.3 & 72.77 & 41.2 & 25.2 & 57.7 & 59.8 & 28.3 & 68.5 & 54.36 \\\hline
Llama-3 3B     & 30.05 & 8.5           & 27.02        & 30.04          & 7.1  & 17.8 & 17.5 & 17   & 83.6         & 26.6  \\
Llama-3 8B     & 68.7  & 69.9          & 63.96        & 90.01          & 58.9 & 92.4 & 81.9 & 62.5 & 91.4         & 75.53 \\ \hline
Qwen-2.5 3B    & 82.2  & 50.5          & 73.17        & 79.9           & 51.6 & 87.5 & 82.1 & 22.4 & 99.9         & 69.92 \\
Qwen-2.5 7B    & 80.5  & 81.1 & 60.36        & 64.9           & 39.4 & 93.3 & 93.9 & 23.3 & 72.3         & 67.66 \\
Qwen-2.5 14B   & 98.5  & 49.7          & \textbf{100} & 99.99 & 86   & 79.4 & 54.1 & 69.6 & 92           & 80.35 \\ \hline
GPT-4o-mini &
  99.1 &
  69.2 &
  99.19 &
  99.7 &
  95.3 &
  99.8 &
  99.2 &
  85.4 &
  95.3 &
  93.57 \\ 
  GPT-4.1 & \textbf{100} & \textbf{98.8} & \textbf{100} & \textbf{100} & \textbf{98.7} & \textbf{100} & \textbf{100} & \textbf{100} & \textbf{100} & \textbf{99.6}
\end{tabular}
\caption{Model performance under the Description + Formalization + In-Context Learning setting across different norms and norm conflicts.}
\label{tab:fg_model_performance_3}
\end{table*}

%% file: appendix/extended_distractors.tex
\begin{table*}[!htbp]
\begin{tabular}{cccccccccc|c}
\hline
\textbf{Model \textbackslash Norm} &
  \textbf{N1} &
  \textbf{N2} &
  \textbf{N3} &
  \textbf{N4} &
  \textbf{N5} &
  \textbf{N6} &
  \textbf{N7} &
  \textbf{C1} &
  \textbf{C2} &
  \textbf{Average Performance} \\ \hline
Granite-3.1 2B &
  11.4 &
  13.3 &
  8.1 &
  9 &
  13.3 &
  31.1 &
  30.3 &
  16.9 &
  30 &
  18.16 \\
Granite-3.1 8B &
  48.2 &
  33 &
  26.72 &
  \textbf{28.1} &
  38.9 &
  35.3 &
  31.6 &
  17.7 &
  41.9 &
  \textbf{33.5} \\ \hline
Phi-3 3.8B &
  44 &
  26.7 &
  21.9 &
  15 &
  14.2 &
  16.9 &
  16.3 &
  12.2 &
  18.2 &
  20.61 \\
Phi-3 14B &
  5 &
  0 &
  0.6 &
  5.1 &
  13.6 &
  27.7 &
  31.2 &
  2 &
  16.9 &
  10.64 \\ \hline
Llama-3 3B &
  0.1 &
  0 &
  0.04 &
  0.3 &
  0 &
  15.9 &
  16.6 &
  8.4 &
  19.5 &
  6.8 \\
Llama-3 8B &
  11.1 &
  12 &
  0.06 &
  1.9 &
  9.9 &
  \textbf{59.7} &
  \textbf{36.5} &
  17.8 &
  21.8 &
  18.95 \\ \hline
Qwen-2.5 3B &
  \textbf{57.1} &
  \textbf{38} &
  \textbf{40.44} &
  18.1 &
  \textbf{45.7} &
  33.8 &
  21.3 &
  15.7 &
  28 &
  33.12 \\
Qwen-2.5 7B &
  29.4 &
  22.3 &
  17.61 &
  11.6 &
  8 &
  25.2 &
  24.7 &
  10.2 &
  30.1 &
  19.91 \\
Qwen-2.5 14B &
  20.8 &
  16 &
  15.21 &
  5.1 &
  21.2 &
  27.8 &
  22.8 &
  \textbf{18.8} &
  \textbf{47.8} &
  21.73
\end{tabular}
\caption{Model performance under the Description Only and Extended Distraction setting across different norms and norm conflicts.}
\label{tab:fg_model_performance_4}
\end{table*}